\documentclass{article}

\usepackage{microtype}
\usepackage{multicol}
\usepackage{graphicx}
\usepackage{subfigure}
\usepackage{booktabs} 
\usepackage{wrapfig}
\usepackage[]{multirow}
\usepackage{float}
\usepackage{amsmath}
\usepackage{dblfloatfix}
\usepackage[export]{adjustbox}
\usepackage{hyperref}

\usepackage[accepted]{icml2021}

\begin{document}

\twocolumn[
\icmltitle{Multimodal Multihop Source Retrieval for Web Question Answering}



\icmlsetsymbol{equal}{*}

\begin{icmlauthorlist}
\icmlauthor{Navya Yarrabelly}{equal,cmu}

\icmlauthor{Saloni Mittal}{equal,cmu}

\end{icmlauthorlist}

\icmlaffiliation{cmu}{Carnegie Mellon University}
\icmlcorrespondingauthor{Navya Yarrabelly}{nyarrabe@andrew.cmu.edu}
 \icmlcorrespondingauthor{Saloni Mittal}{salonim@andrew.cmu.edu}

\vskip 0.3in
]




\begin{abstract}
This work deals with the challenge of learning and reasoning over multi-modal multi-hop question answering (QA). We propose a graph reasoning
network based on the semantic structure of
the sentences to learn multi-source  reasoning paths and find the supporting facts across both image and text modalities for answering the question.  In this paper, we
investigate  the importance of  graph structure for multi-modal  multi-hop question answering. Our analysis is centered on WebQA. We construct a strong  baseline model, that finds relevant sources using a pairwise classification task. We establish that, with the proper use of feature representations from pre-trained models, graph structure helps in improving multi-modal
multi-hop question answering. We point out
that both graph structure and adjacency matrix
are task-related prior knowledge, and graph structure can be leveraged to improve the retrieval performance for the task. Experiments and visualized
analysis demonstrate that message propagation over graph networks  or
the entire graph structure can  replace massive
multimodal transformers with token-wise cross-attention. We  demonstrated the applicability of our method and show a performance gain of \textbf{4.6$\%$}  retrieval F1score over the transformer baselines, despite being a very light model. We further  demonstrated the applicability of our model to a large scale retrieval setting.
 
\end{abstract}

\section{Introduction}
\label{introduction}

\begin{figure}[t]
    \centering
    \includegraphics[scale=0.25]{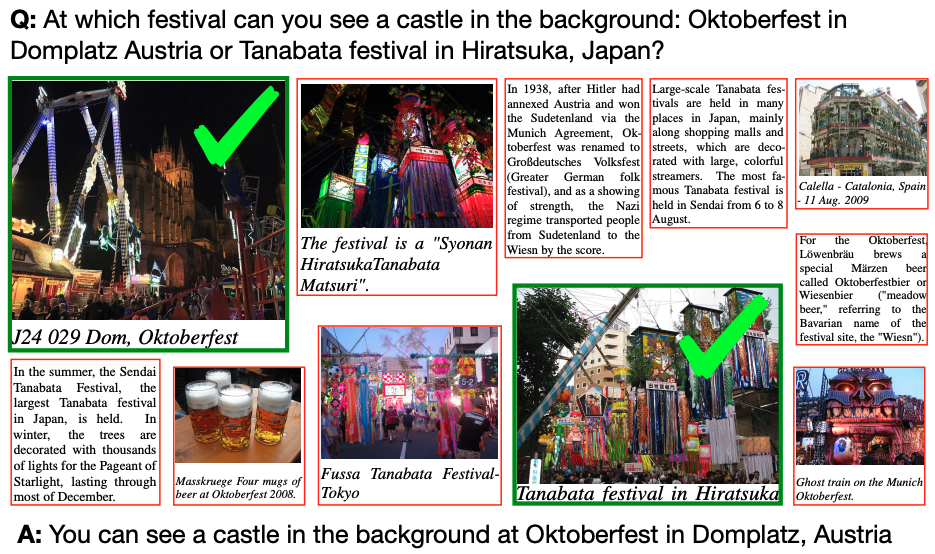}
    \vskip -0.1in
    \caption{Top: Sample query; Mid: Possible Sources; Bottom: Desired response}
    \label{Fig_1}
    \vskip -0.2in
\end{figure}

Web search is fundamentally multimodal and multihop. Here multimodal refers to the fact that the web has text, images, tabular and speech modality data, and when we query, the response should be based on the features extracted from all these modalities. However, most web search engines treat the web as a text-only landscape which results in not leveraging many highly informative sources. 
In general, when humans search the web, they go through multiple sources, and effectively aggregate information based on reasoning to answer complex questions . So, essentially we need data that is multihop - a question requires reasoning over information scattered over multiple sources in order to obtain the correct answer.
In this project, we aim to target the problem of multimodal multihop source retrieval for open-domain question answering as shown in Fig. \ref{Fig_1}. Specifically, we aim to develop an AI system which for a given text question, can select relevant sources of information from different modalities required to generate a correct answer. 

Graph Convolutional Networks (GCNs) are designed to share information across different nodes of a graph and effectively utilize them to make decisions. They are hence well suited for the task of multihop reasoning. However, to the best of our knowledge, there is no existing work that aims to explore GCNs for solving multimodal multihop source retrieval. 
In this work, we explore independent approaches each one designed to explore different aspects of GCNs and showcase their impact on our specific task. Through this work, we demonstrate the comparable performance of GCNs to complex transformer-based baselines~\cite{WebQA21} even with lightweight sub-optimal input representations. We suspect this is critically due to the inductive biases of GCNs making them well suited for information aggregation and multimodal retrieval tasks.

The primary contributions of this work are as follows:
\begin{itemize}
    \item A novel graph-based framework to solve multimodal and multihop source retrieval that can scale to open-domain question-answering in the wild.
    \item Generate question-conditioned node representations of sources that are informed by the information present in other sources.
    \item Explore various graph architectures for the source retrieval problem.
    \item Present the computation and latency advantages of our proposed approach over expensive transformer-based models which makes it scalable for a web-scale retrieval task.
\end{itemize}


This is the link to our Github repository for the project: \url{https://github.com/smittal10/Multimodal-Multihop-QA}
\section{Related Work}
\label{related work}
\subsection{multimodal visual Q/A}
The seminal work of \cite{VQA} released a large-scale dataset for Visual Question Answering. This work extracted uni-modal deep features and fused (point-wise multiplication) them for answering. \cite{visdial} introduced visual dialog task. The aim of this problem is to develop a conversational AI agent which can also process inputs from visual modality. However, there is a gap between the quantitative (reported) performance and the actual ability of models to generate realistic and diverse inputs. 

The work from \cite{visdial_diversity} aims to improve the generative ability of models for real-life scenarios. They introduce two competing agents Q-BOT and A-BOT. Q-BOT is trained to ask diverse questions which forces A-BOT to explore a larger state space and jointly answer more informatively. The work of \cite{balanced_vqa_v2} was aimed to balance the VQA dataset by having almost twice the original image-question pairs. The pairs often contained conflicting question-answers for the same visual cues to prevent overfitting. Interestingly, extensive works still make a weak assumption that all relevant information is often limited to a single image. Although the context needed to answer many real-life questions can span across multiple images.




\subsection{Multihop QA}
There has been substantial work in recent years on building QA models that can reason over multiple sources of evidence. In 2018, \cite{Yang2018HotpotQAAD} introduced a text-based QA dataset, HotPotQA that required reasoning over multiple supporting documents. They used an RNN-based architecture that could produce “yes”/“no” or span-based answers. This seminal work is still uni-modal in nature.

Recent works have expanded multihop QA for multimodal inputs. The work from \cite{Talmor2021MultiModalQACQ} proposed MultiModalQA, which uses inputs image, text, and table to answer questions. They propose a multihop decomposition (ImplicitDecomp) method where the answer is iteratively generated by parsing through individual QA modules. However, they generate questions from a fixed template and hence task reduces to filling of missing entries once the template is identified. Hence it cannot scale well for zero-shot generations. MIMOQ \cite{Singh2021MIMOQAMI} proposes a QA system that can reason and also respond in multiple modalities. The authors propose a novel multimodal framework called MExBERT (Multimodal Extractive BERT) that uses joint attention over input textual and visual streams for extracting multimodal answers given a question. They observe noticeable improvements when compared with methods that independently extract answers from unimodal QA modules. However, the key contribution of MIMOQ is still to generate multimodal answers and not information aggregation across modalities.

\subsection{Cross modality representations}
Significant advances in Visual QA and multimodal QA can be attributed to the advances in Transformer-based methods initially proposed by \cite{10.5555/3295222.3295349} and their large-scale pre-training methods. There are two major directions of pre-training: (a) Parallel streams of encoders one for each modality followed by fusion \cite{LXMERT}\cite{VILBERT}; (b) Unified encoder-decoder representations that can take both the language or image modalities \cite{VLP_PAPER}. \cite{ANDERSON} proposed a new form of representation for images. They extract objects from the image and represent an image as a set of embeddings for the Regions of Interest (ROI). The spatial positional features (bounding box locations) are also added to the encoding. This representation is semantically closer to language representations. Hence refining these models and ways of representations for multihop reasoning is expected to work well on WebQA \cite{WebQA21} benchmark.

To the best of our knowledge, this is the first dataset that introduces open-domain question-answering in a multimodal and multihop setting. Success on WEBQA requires a system to retrieve relevant multimodal sources first and aggregate information from text and vision modality.
Our approach significantly differs from all the prior work in this domain, where we leverage GNNs to solve the problem of multimodal retrieval and ranking. One major limitation of transformer-based models is that they cannot process all the sources together while making a decision during retrieval as the input length is limited. In an open-domain QA system that typically has hundreds of candidate sources to choose from, this problem is amplified manifold. Our approach solves this fundamental problem as it can leverage information from all the sources at the same time.

\section{Problem Statement}
\label{problem statement}
For multimodal multihop QA the problem is defined such that the QA system gets a text question $Q$ and a set of sources $s_i \in \mathcal{S}$ as input. It is important to note that source $s_i$ can be either an image $\mathcal{I}_i$ or a  text snippet $\mathcal{T}_i$ such that $\mathcal{S} \subseteq \mathcal{I} \cup \mathcal{T}$. The benchmark in \cite{WebQA21} divides the problem into two separate tasks:

Task (A) is a source separation task. Here, for a given question $Q$ system has to divide the sources $\mathcal{S}_Q$ into positive and distractors (${\mathcal{S}_{Q}^{+},\mathcal{S}_{Q}^{-}}$). Hence formally the problem is defined as :
\begin{equation}
    \mathcal{S}_{Q}^{+} = f_{source} (Q,\mathcal{S}_Q | \phi) \:\:\: \text{where } \mathcal{S}_{Q}^{+} \subseteq \mathcal{S}_{Q} \subseteq \mathcal{S}
    \label{eq:2}
\end{equation}
Task (B) is the Question Answering task. Here for a given Q and $\mathcal{S}_{Q}^{+}$ the task is to learn a function to generate $A_{pred}$ such that
\begin{equation}
    A_{pred} = f_{qa} (Q, \mathcal{S}_{Q}^{+} | \theta)
    \label{eq:3}
\end{equation}
$A_{pred}$ is the generated answer (not classification). The functions $f_{qa}$ and $f_{source}$ are parameterized by $\phi$ and $\theta$ which are the weights of neural models. This project primarily focus on Task (A) i.e. the problem of multimodal multihop source retrieval.

\section{Baseline Models}

\subsection{VLP + VinVL}
The authors from WebQA~\cite{WebQA21} introduced this baseline for multimodal source retrieval. VLP~\cite{VLP_PAPER} is a pre-trained multimodal transformer well suited for both understanding and generative tasks. The source retrieval baseline is fine-tuned on the VLP checkpoints. Text modality inputs are tokenized by Bert-base-cased tokenizers. For images, the baseline extracts $\sim 100$ region proposals using latest state-of-the-art model VinVL~\cite{zhang2021vinvl}. The authors conducted experiments with various image encoders like x101fpn etc. however demonstrated VinVL~\cite{zhang2021vinvl} to work best. In this work, we hence draw comparisons with the baseline's best VLP + VinVL architecture.

For a given question $Q$, every source $\mathcal{S}_{Q,i} \in \mathcal{S}_Q$ is fed into the baseline model one by one. The input to the model is hence a pair of (Source, Question) i.e. $<[CLS], \mathcal{S}_{Q,i}, [SEP], Q, [SEP]>$ and the model predicts the probability of it being a positive source. This approach hence makes a critical assumption that prediction over a source is independent of other sources in the dataset. We find this to be a weak assumption and not in the spirit of multihop reasoning which the dataset~\cite{WebQA21} critically demands.    

\subsection{CLIP + Sentence-BERT based baseline}
\label{4.2}
As mentioned in the above section, the VLP baseline introduced by the authors uses very expensive input features. For example, the image features for WebQna dataset alone account for 500 GB of disk space (as mentioned by the author in thier official github repo ~\footnote{\url{https://github.com/WebQnA/WebQA_Baseline}}) as it takes in 100 region proposals for every image. For our proposed method, we use very lightweight input features (discussed in the later sections) from pre-trained models like CLIP \cite{Radford2021LearningTV} and Sentence BERT  \cite{Reimers2019SentenceBERTSE} which fit under 2 GB of memory. In the spirit of fair comparison, we train a baseline model with the same pairwise classification objective as the VLP baseline keeping the input feature space same with our proposed graph-based model. The motivation to build this baseline is to determine whether a graph-based approach has merit over a pairwise objective (as in VLP baseline) keeping the input feature space same. For this purpose, we encode the text modality (question, text snippets and image captions) with Sentence-BERT (sBERT) and images with CLIP's vision encoder. For a question-source pair we obtain encoded vectors after passing through CLIP and sBERT. We apply a few linear layers on top of concatenation of question-source encoded vector and predict the probability of it being a positive source for that question. We fine-tune the weights of CLIP and sBERT while training on this task.

\section{Proposed Approach}

\section{Methods}
Our proposed Hierarchical
Graph Network (HGN) consists of four main components: (i) Graph Construction Module
to construct a hierarchical graph to include connections from questions to sources and  edges to connect question guided knowledge extraction from different sources; (ii) Context Encoding Module, where initial
representations of graph nodes are obtained via various pre-trained transformer models for both images and textual sources ; (iii) Graph Reasoning
Module, where graph-convolution-based
message passing algorithm is applied to jointly
update node representations; and (iv) Graph
Prediction Module, where we combine different training objectives to train out graph-encoder modules to retrieve relevant sources for the given question.
We investigated different architectures and graph structures and training objectives to identify  the best combination for effectively identifying the relevant sources  retrieval for the problem of multi-modal multi-hop question answering. We briefly describe each of the components and include the empirical analysis in the section below. 
\begin{figure}
    \centering
    \includegraphics[width=\linewidth]{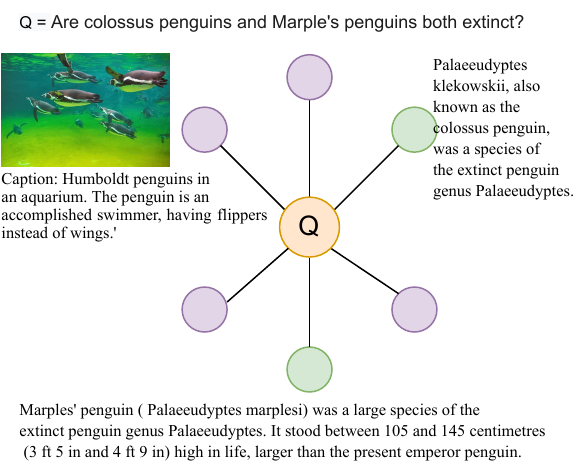}
    \caption{We disentangle the question node $Q$ in yellow. There are considerably less edges in this architecture. This example suggests positive text snippets containing relevant information}
    \label{fig:star}
\end{figure}

\subsection{Graph Construction Module}
\subsubsection{Star Node Structure}

Figure \ref{fig:star} shows the graph structure where each source is only connected to the question and the information aggregation across source happens in a  multi-hop setting 
ion to all source node. For this approach, all relevant information to identify a particular source as positive is within two hops. 

\subsubsection{Fully Connected Graph Structure}
Figure \ref{fig:hgn_entity} shows the fully connected strcuture where we add additional edges between all the source nodes. This structures enables for the model to enable direct communication between the sources but has the risk of aggregating irrelevant information from different sources.

\subsubsection {Hierarchical Semantic Graph Networks}
\begin{figure*}
    \centering
    \includegraphics[scale=0.6]{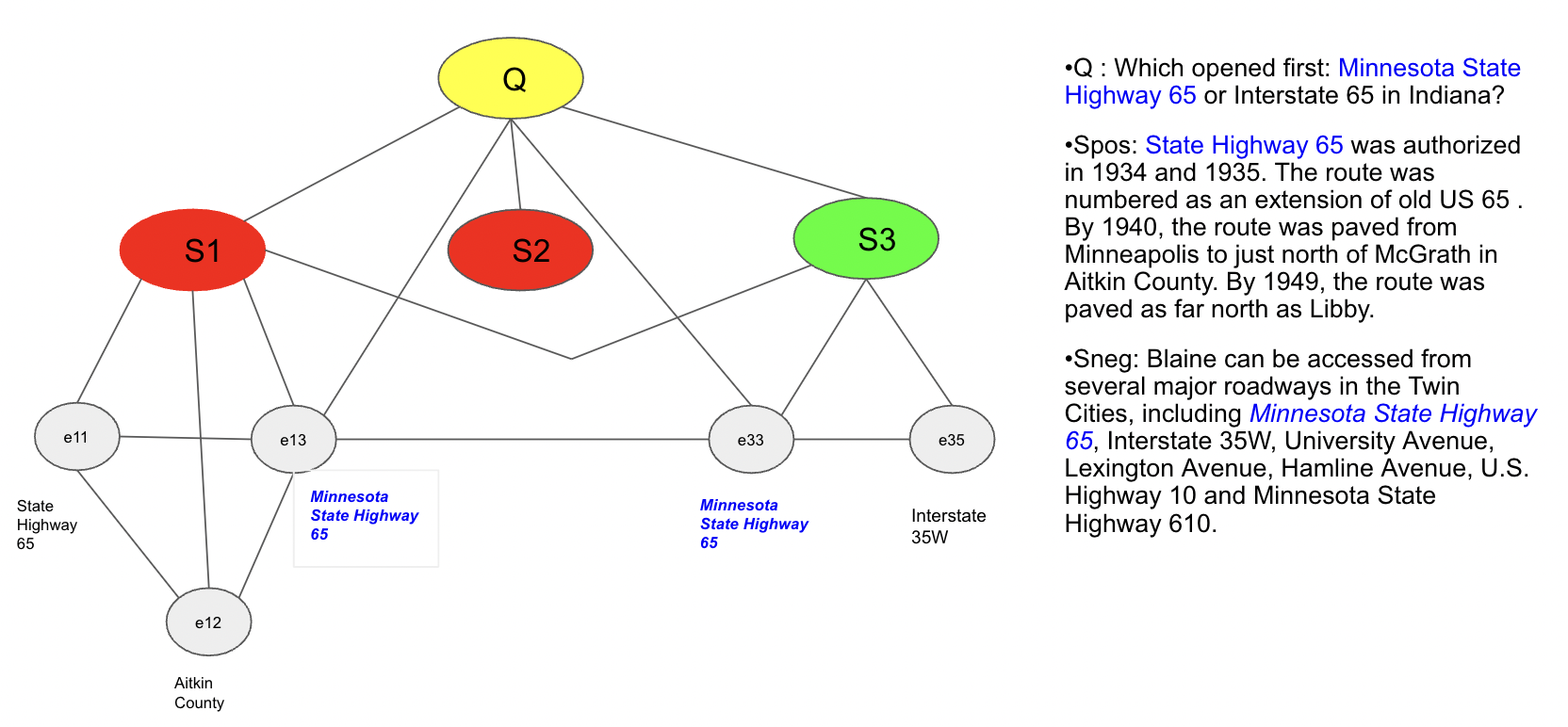}
    \caption{Entity based Hierarchical Graph Network. Nodes in yellow, represent Questions, red denotes distractor sources and green indicates postive sources and nodes in grey are the entity nodes for each of the source and question nodes. }
    \label{fig:hgn_entity}
\end{figure*}

\textbf{Entity based GNN}  
\newline
We rely on textual part of both images and text sources to further build a fine-grained heirarchical network for our task. Each question and source are linked via the entities that are being addressed.  Thus our graph is naturally encoded in a hierarchical structure, and  motivates our graph  construction based on 
the hierarchical connections across sources and source to question.
For each source node,
we add edges between the node and all the sentences in the source. We further extract all the entities in the sentence and add
edges between the sentence node and these entity
nodes. Each question also has entities and the entities across different sources and question are connected if the mention refers to the same entity. 
Each type of these nodes captures semantics
from different information sources. Thus, the hierarchical graph effectively exploits the structural
information across all different granularity levels
to learn fine-grained representations, which can locate supporting facts and answers more accurately
than simpler graphs with homogeneous nodes. Figure ~\ref{fig:hgn_entity} describes the construction of the heterogeneous entity graph. The entity "Minnesota State Highway" is refered in both sources S\textsubscript{1} , S\textsubscript{3} and Question Q. Thus, we  form edges between S\textsubscript{1} and  S\textsubscript{3} and also between each of the entity mentions across different sources. Since each source captures its own contextual representation of the shared entities, by creating edges we enable for our graph reasoning modules to allow for an information flow  path from question to sources and between sources to answer multi-hop queries \cite{fang2019hierarchical}.
\begin{figure}
    \centering
    \includegraphics[scale=0.4]{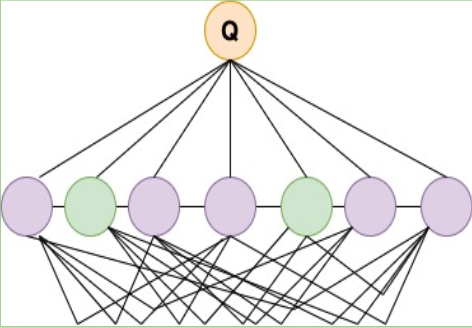}
    \caption{Here we build upon the star-based architecture by adding source-source dense connections. Full connections between sources ensure that information relevant to make decision for a node is available in a
single hop. }
    \label{fig:star}
\end{figure}

\textbf{Semantic Role Labelling based GNN} \newline
Incorporating semantic informnation has shown to improve the performance of multihop question answering systems \cite{zheng2020srlgrn}. 
We further experimented to inlcude the semantic structure of the source to  build a heterogeneous graph that contains document-level sub-graph S and argument predicate SRL sub-graph Arg for each data instance. In the graph construction process, the
document level sub-graph S includes question q, and sources $ S_{1}^{1}, \ldots, n$ which includes both textual and image modalities. The
argument-predicate SRL sub-graphs Arg, including arguments as nodes and the predicates as edges,
are generated using AllenNLP-SRL model. We extract the arguments corresponding to each predicate and create a node which is a concatenation of the  argument phrase and argument type,including “TEMPORAL”, “LOC”, etc.
Figure ~\ref{fig:hgn_entity} describes the construction of the heterogeneous graph. The heterogeneous graph’s edges
are added as follows: 1) There will be an edge between a source and an argument if an argument
appears in this source  2) Two sentences s\textsubscript{i} and s\textsubscript{j} are connected by an edge if they share an argument by approximate textual matching (the red dashed lines); 3) Two argument nodes
Arg\textsubscript{i}
and Arg\textsubscript{j} will have an edge if a predicate exists between Arg\textsubscript{i}
and Arg\textsubscript{j} 4) There will be an edge between the question and
source if they share an argument. 
\par 
We build a predicate-based semantic edge matrix K and a heterogeneous edge weight matrix A. The semantic
edge matrix K is a matrix that stores the word index of the predicates. If two argument nodes Arg\textsubscript{i}
and Arg\textsubscript{j} 
related to the same predicate, we add
that predicate word index to K(Argi
, Argj). Sometimes, Arg\textsubscript{i}
and Arg\textsubscript{j} 
are related to more than one
predicate. These predicate features can be used as edge features for various graph encoding modules. 
In the meantime, the heterogeneous edge weight.  matrix A is a matrix that stores different types
of edge weights. We divide the edges into
three types: source-argument edges, argument-argument edges, question-source, question-argument and source-source edges. The weight of a sentence-sentence edge is 1
when two sentences share an argument. Meanwhile, the weight of a sentence-argument edge is
1 if there exists an edge between a sentence and
an argument. The edge weight matrix can  also be modified to include the prior knowledge of relevance of sources to a given question.
\par
\textbf{Node Representations} 
For each of the sources $ S_{1}^{1}, \ldots, n$, we initialize the node features for textual sources with a 768 dimensional BERT embedding and for image sources, we concatenate the image representations obtained using different pre-trained models and the caption embedding. Since each argument occurs in multiple sources, each of these arguments are initialized with their phrase embeddings obtained from the corresponding source. This allows for different sources to exchange information about 

\subsection{Context Encoding Module}
To encode each modality we use modality-specific encoders. All the text information i.e. text snippets, image captions, and questions are encoded into a 768-dimensional vector using sBERT \cite{Reimers2019SentenceBERTSE}. We used the pre-trained variant of Sentence BERT on MSMARCO Passage Ranking Dataset~\footnote{\url{https://www.sbert.net/docs/pretrained_models.html\#msmarco-passage-models}}. Please note that by using such an encoding scheme we lose token-level features and embed the entire document/sentence in one vector. This can be seen as a downside of our approach over the baseline VLP model that takes token-level features and uses the attention mechanism on them.

    To encode the vision modality, we experiment with two pre-trained models; (i) ResNet-152 \cite{He2015} network - 2048 dimensional vector and (ii) CLIP-ViT-L/14 \cite{Radford2021LearningTV} - 768 dimensional vector. ResNet has a CNN-based model architecture and is pre-trained on an image recognition task. CLIP is a multi-modal model that is trained on millions of text-image pairs on a contrastive similarity objective. While ResNet is a unimodal encoder, the Vision Transformer in CLIP is tuned towards jointly understanding text and images.
We experiment with these input encoders in the following settings:
\begin{enumerate}
    \item Both text and vision modality encoders used as feature extractors in a zero-shot fashion.
    \item Jointly fine-tuned vision encoder (specifically, CLIP-ViT-L/14) with GNN module training.
    \item Used the fine-tuned CLIP and sBERT modules from baseline training \ref{4.2} as feature extractors during GNN module training.
\end{enumerate}
\subsection {Graph Reasoning Module}
We construct a multimodal heterogenous graph where a node either represents a question or a source. The edges between these nodes are added based on the graph strutures explained in the previous section.

To learn the graph embedding we leverage from the learning of GraphSAGE~\cite{10.5555/3294771.3294869}. This work proposed a method to inductively learn node embeddings and the method generalizes well to previously unseen data. Specifically we use the following function to aggregate information across nodes:
\begin{equation}
\label{eq:sage}
    x^{\prime}_i = W_1 x_i + W_2 \cdot mean_{j \in \mathcal{N}(i)} x_j
\end{equation}
Here $\mathcal{N}(i)$ denotes the 1 hop-neighborhood of node with index $i$.
We use multiple layers of GraphSAGE that facilitates information aggregation from the distant neighbors as well. We train the GraphSAGE module with different training objectives described in the next section. When the graph converges, we obtain multi-modal node embeddings that map image and text sources in the same space.
\subsection{Graph Training Objectives}

\subsubsection{Node Classification}
Our Graph Encoder $G_{p}(.)$ module maps each source and question to a $d$ dimension real valued vector.  
Each question node $q_{i}$ is connected to the positive sources $s_{i, 1}^{+}$  and negative sources $s_{i, 1}^{-}$. Hence, the transformed embedding of each node is a contextual representation with question knowledge encoded in it. We aimed to learn whether each source is a positive or negative source from its embedding vector. We train an MLP on top of the graph encoder module to predict the binary class of each source and used a binary cross entropy based loss function for all the nodes for a question.
 
\subsubsection{Edge Classification}
To further guide the Graph Encoder module, we also tried to explicitly predict the edge connection type  (whether an edge exists or not) between question and the sources and also between sources to reinforce the multihop connections signal among sources to the Graph Encoder training. We use cosine similarity of the transformed node embeddings to  determine the probability of an edge between two nodes and use a binary cross entropy based loss function for each edge in the graph.

\subsubsection{Contrastive loss}
Since we construct a heterogeneous graph encoding module, the embeddings of nodes of different types are transformed using different feature transformation but in the same vector space. Dot-product similarity as a  ranking function to train the the graph encoders for the problem of retrieval is considered a metric learning problem ~\cite{kulis2013metric}. The objective is to embed all the nodes in a vector space where the similarity between a question and a positive passage is higher than the question and it's corresponding  dis-tractor sources and we aim to do this by learning a better embedding space. 

Training the graph encoders so that the dot-product similarity (Eq. (1)) becomes a good ranking function
for retrieval is essentially a metric learning problem (Kulis, 2013). The goal is to create a vector
space such that relevant pairs of questions and passages will have smaller distance (i.e., higher similarity) than the irrelevant ones, by learning a better
embedding function. 
\newline
Let $\mathcal{D}=\left\{\left\langle q_{i}, s_{i}^{+}, s_{i, 1}^{-}, \cdots, p_{i, n}^{-}\right\rangle\right\}_{i=1}^{m}$ be a training data that contains a question $q_{i}$ and positive sources $s_{i, 1}^{+}$ and distractor sources $s_{i, 1}^{-}$. We optimize the loss function as the loss function as the negative log likelihood of the positive sources. 

\begin{equation}
\begin{array}{c}
L\left(q_{i}, s_{i}^{+}, s_{i, 1}^{-}, \cdots, s_{i, n}^{-}\right) \\
=-\log \frac{e^{\operatorname{sim}\left(q_{i}, s_{i}^{+}\right)}}{e^{\operatorname{sim}\left(q_{i}, s_{i}^{+}\right)}+\sum_{j=1}^{n} e^{\operatorname{sim}\left(q_{i}, s_{i, j}^{-}\right)}}
\end{array}
\end{equation}

\section{Experimental Setup}
\label{methodology}
\subsection{Dataset}
We use the WebQA \cite{WebQA21} dataset to solve the problem of multimodal multihop source retrieval. The dataset contains input from two different modalities: \textit{Images} and \textit{Text}.  Table \ref{tab: modality statistics} summarises the number of instances of textual and visual source-based queries in the train, val, and test dataset. Each question $Q$ is text-based and the dataset contains a set of sources $\mathcal{S}_Q$ associated with each question. Each element in the set $\mathcal{S}_Q$ can either be a positive source (with label $+1$) or a negative source (with label $0$). Note that the number of sources ($|\mathcal{S}_Q|$) is not fixed and hence is different for each question $Q$. The questions with a positive source from visual modality are further sub-divided into six categories. Table \ref{tab: question_categories} contains these question categories. We do not specifically use this sub-division while training however we do use them for quantitative analysis of performance.
\begin{table}[t]
    \centering
    \vline
    \begin{tabular}{c|c|c|c|}
        \hline
        \textbf{Modality} & \textbf{Train} & \textbf{Dev} & \textbf{Test}\\
        \hline
        Image &  18,954 & 2,511 & 3,464\\
        \hline
        Text & 3,464 & 2,455 & 4,076\\
    \hline
    \end{tabular}
    \caption{Statistics by the modality required to answer the question.}
    \label{tab: modality statistics}
\end{table}

\subsection{Evaluation Metrics}
The performance for source retrieval is measured using the F1-score.
\subsubsection{F1-Score}
There is a severe imbalance in the dataset and hence accuracy of source selection is not the best metric. For example, the base model can achieve $\sim92\%$ accuracy by predicting every source as negative. Hence we use the F1-score to measure performance. F1-score is measured as a harmonic mean of Precision and Recall.
\begin{equation}
\label{eq:fscore}
    F1\-score = 2 * \frac{Pr * Recall}{Pr + Recall}
\end{equation}
where Precision ($Pr$) is defined as $\frac{TP}{TP + FP}$ and $Recall$ is defined as $\frac{TP}{TP + FN}$. Further, $TP$ are the true positives, $FP$ are false positives and $FN$ are false negatives. F-score enables us to measure of well the model is in predicting positive and negative sources even with imbalance. 
\subsection{Experiment details}
Section 4 describes our baseline models. Our baselines operate in a pairwise setting, they take a question and a source and predict the relevance of each source individually,  as the transformer based architectures  cannot handle large input lengths. 



\begin{table*}[h]
\centering
\begin{adjustbox}{width=0.9\linewidth,center}
\begin{tabular}{l cccccc }
    \toprule
    No. & Model & Features & F1 & Text-F1 &  Image-F1 \\
    \midrule
    1 & Baseline -  Pairwise VLP & 100 Region features +  BERT token feat &  68.9 & 69.48 &  68.13  \\
    \midrule
     2 & Pairwise(baseline) & (zero-shot) Resnet152 + sBERT & 48.7 & - & -\\
    3 &  Pairwise (baseline) &  (fine-tuned) CLIP + sBERT & 56.3 & - & -\\
    4 &  GNN - Fully Connected  &  (zero-shot) Resnet152 + sBERT  & 60.8  & - & -\\
    5 &  GNN - starnode  &  (zero-shot) Resnet152 + sBERT  & 62.8  & 58.73 & 63.66\\
    6 &  HGNN with Named Entities  &  (zero-shot) Resnet152 + sBERT  & 62.3 & 63.05 &  62.1 \\
    7 &  HGNN with SRL  &  (zero-shot) Resnet152 + sBERT  & 64.8 & \textbf{67.3} & 63.9 \\
    8 &  GNN - starnode   & (zero-shot) CLIP + sBERT  & 65.8  & 61.38 & 69.84 \\
    9 &   GNN - starnode   & CLIP (joint fine-tuning) + sBERT(zero-shot) & 64.4 & 61.04  &  67.2\\
    9 &   GNN - starnode   & CLIP (fine-tuned) + sBERT(fine-tuned) & \textbf{67.4} & 61.9  & \textbf{72.73}\\
    \bottomrule
\end{tabular}
\end{adjustbox}
\caption{The tables lists the source retrieval F1score of baselines and our methods on the WebQA dataset.}
\label{table:results_gnn_all}
\end{table*} 

\section{Results and Discussion}

Table ~\ref{table:results_gnn_all} lists our main results on the WebQA dataset in a restricted setting where we rank only 50 relevant sources for each question. Row 1-3 lists the baselines which are all pairwise classification models that take a question and candidate source and determine the relevance of each source independently. VLP baseline (row 1) uses a multi-modal transformer with token level cross-attention between the modalities by taking 100 region proposal features for an image and BERT features for representing question and passage tokens. Since these features are very expensive , we constructed two additional baselines that use CLIP\cite{Radford2021LearningTV} for image features and Sentence-BERT\cite{Reimers2019SentenceBERTSE}  for textual features. Our GNN with star graph using fine-tuned CLIP and fine-tuned sBERT features gave the best combined F1 among all the settings and about +4.6\% increment over the pairwise VLP baseline for image queries. This shows the efficacy of our approach despite using much cheaper features (source level features instead of token level) and a much lighter model that can rank all the sources in a single pass. 
\par
Table ~\ref{tab: question_categories} shows F1 score for different categories of queries. Despite smaller  number of training examples for the categires "shape" and "others" , we are able to achieve an F1 score comparable with that of other categories with large number of traning instances available. 

\subsection{Graph Structures Analysis}
We experimented with three kinds of graph structures. 1. Starnode structure where we include connections from question to all source node. 2. Fully Connected structure where we add additional connections between all source nodes. 3. Hierarchical graph  networks as discussed in section 6.1. Row 5 refers to the fully connected graph structure and rows 5,8,9 refer to the F1 score results with the starnode structure and rows 6 and 7 refer to the Hierarchical Graph  Networks (HGN). We observed that adding additional negative connections among source nodes leads to irrelevant information aggregation and since the GCN model cannot dynamically determine the edge weights, fully connected graph structure led to a decreased performance. Overall we saw an improvement of about 2.3 \% overall gain in F1 score and about 8.7\% increase in the F1 score for the text queries. 
\par

From our fully connected vs starnode results analysis, we believe that Graph attention networks which can dynamically adjust the weights for the neighbors for query dependent source aggregation would further improve the performance. This establishes that having fine-grained nodes and incorporating a rich semantic structure helps in the retrieval performance. We believe that having a richer input features and more fine-grained nodes for images would have improved for image queries also. We plan to further explore this direction for improving image-retrieval accuracy.  
\subsection {Context Encoding Module -Features}
\label{Context_Encoding_Module}
Table \ref{table:results_gnn_all} shows the impact of our different modality-specific features input to our GNN model. From rows 6 and 8, it is evident that zero-shot CLIP features for images is more informative to graph learning than zero-shot ResNet features. We get an improvement of +3 points in combined F1 score by using CLIP features as we expected earlier since CLIP is trained on a multimodal objective. Using CLIP features not only improves the Image F1 but also improves Text F1 by +2.65 points. We believe this happens because of the better text-aligned image features from CLIP that makes the model discard the negative image sources relatively easier now. Jointly fine-tuning CLIP's vision encoder with the GNN module results in degradation of performance by -1.4 points. This result aligns with one of the conclusions in this work \cite{Shao2020IsGS} where authors say that the graph structure is helpful when pre-trained models are used in a feature-based manner and not jointly fine-tuned. In row 9 of table \ref{table:results_gnn_all}, we report the results for GNN model with image and text features extracted from fine-tuned CLIP and Sentence BERT respectively from our second baseline model (row 3). This is the best model that we attain in all our setting giving a boost of +1.6 points in F1 over zero-shot CLIP and sentence BERT features. \textbf {It also helps to give a massive improvement in image queries over the baseline by +4.6 points}.
\begin{figure}
    \centering
    \includegraphics[scale=0.5]{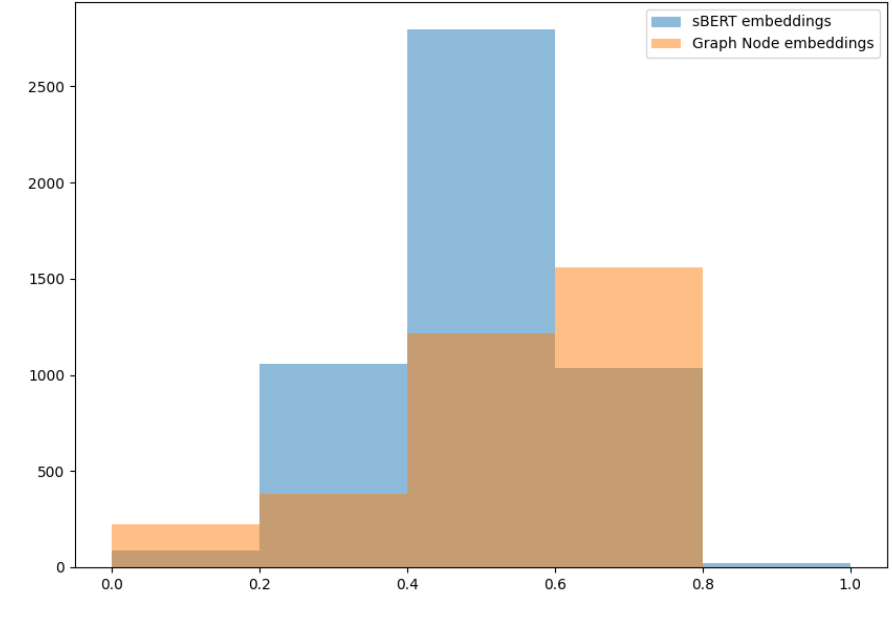}
    \includegraphics[scale=0.5]{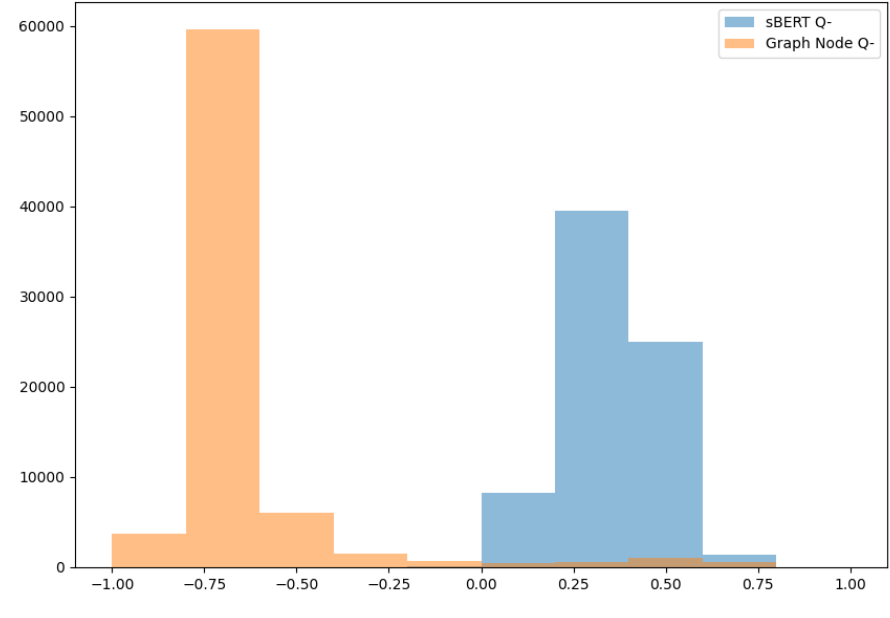}
    \caption{The above two plots show the dot-product distribution between Question and source embeddings. The above graph represents the distribution of Question-Negative source pairs, while the graph below represents the distribution for Question-positive source pairs. The histogram in blue is for similarities obtained from using pre-trained sBERT embeddings and in orange is using graph node embeddings}
    \label{fig:qsrc_sim}
\end{figure}
\subsection{Question-Source similarity Visualizations}
Through the histogram plots in figure \ref{fig:qsrc_sim} we wish to show the power of our graph-based method (in orange) in learning effective representations for question and sources over pre-trained models like sentence BERT (in blue). The above and below graphs shows the dot product distribution of (question (.) negative text source) pairs and (question (.) postive text source) pairs respectively. By plotting the dot product, we want to compute a similarity score between question and source. A good embedding representation would show high similarity between question and positive sources (close to 1), while low similarity between question and negative sources (close to 0). We can see in the above graph how the similarity scores obtained from graph node embeddings (in orange) have the majority chunk between 0.6 and 0.8 while the Sentence BERT-based similarity scores peak in the range of 0.4-0.6. 
Similarly, in the graph below we observe the efficacy of graph node-based embeddings (orange) as the distibution of question and negative source similarity scores move below 0 close to -0.75 while sentence-BERT based scores mainly lie between 0.25 and 0.6.
\begin{figure*}
    \centering
    \includegraphics[width=0.9\linewidth,height=8cm]{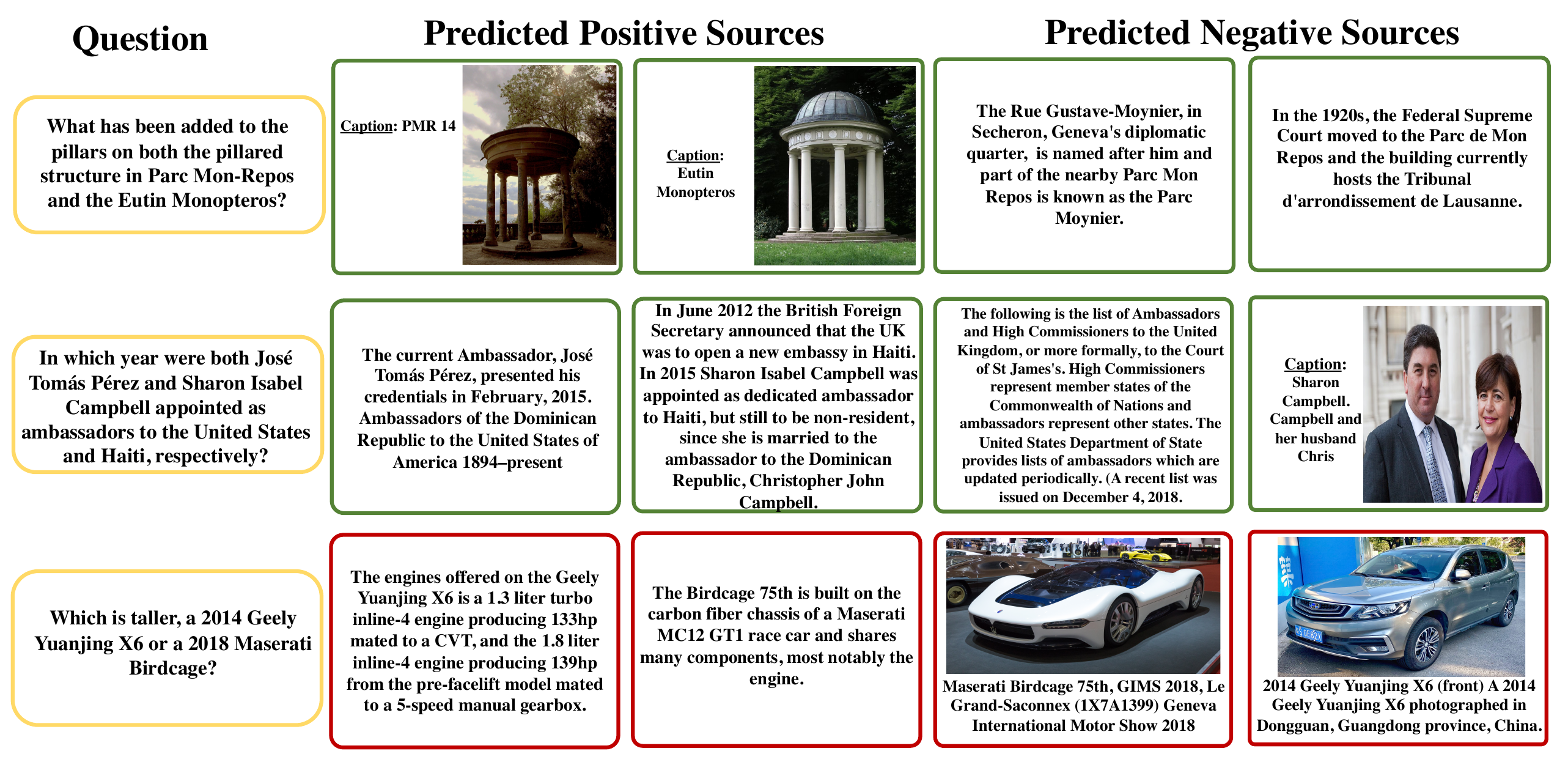}
    \caption{Retrieval Results for our GNN approach: Green box indicate predictions which were correct while red indicates incorrect}
    \label{fig:star}
\end{figure*}
\subsection{Retrieval in a Restricted vs Full Setting}

For handling web scale retrieval, we further investigate the effect
of retrieval scale where the QA system has to rank all the available sources (of order millions) to determine the positive sources that contain the information to answer a given query. For a pairwise classification baseline like VLP ~\cite{WebQA21}, it would take 3 years to rank 1 Million sources for a given query and hence such a model cannot work for full scale retrieval. We measure the performance impact by filtering top 20 sources using inexpensive similarity metrics and re-ranking the top-20 sources using our models. \par

Table ~\ref{tab: full_scale} shows the results for dense (BM25) versus sparse retrieval approaches. For the VLPbased model, top 20 sources are retrieved based on CLIP embedding similarity and then these sources are re-ranked using VLP baseline, sources
are selected if its binary classification confidence is above a specified threshold. For our GNN based model, we follow a similar procedure to filter top 20 sources based on sBERT similarity score and re-rank them using our GNN model. We observed that while sparse retrieval approaches worked well for restricted setting, the performance dropped for an increased scale.  While our GNN model could have handled larger number of candidate sources, it suffers from over squashing from negative sources. However,  it holds the promise for Graph Attention Networks which can prune negative edges more efficiently.

\begin{table}[b]
    \centering
    \vline
    \begin{tabular}{c|c|c|c}
        \hline
        \textbf{Model} & Img-F1 & Txt-F1 \\
        \hline
        BM25 & 20.43  & 28.15 \\
        \hline
        CLIP(20) + VLP & 21.68 & 26.01 \\ 
        \hline
        SBERT(20) + GNN (ours)  &  22.84 & 24.58 \\
    \hline
    \end{tabular}
    \caption{F1 accuracy for different models for a fullscale retrieval setting}
    \label{tab: full_scale}
\end{table}
\begin{table}[b]
    \centering
    \vline
    \begin{tabular}{c|c|c|c}
        \hline
        \textbf{Model} & \#Num Sources & Retrieval Time \\
        \hline
        VLP (baseline) & 50  & 250 milliseconds\\
        \hline
        GNN(Ours) & 50 & 1 millisecond \\ 
        \hline
    \hline
    \end{tabular}
    \caption{Latency analysis for retrieval problem on WebQA dataset}
    \label{tab:latency}
\end{table}

\subsection {Computation and Latency Analysis}
We also analysed the computation requirements of our model vs the baseline transformer based VLP model. Table ~\ref{tab:latency} shows the latency to process  50 sources for a question for each of the models. Our GNN based model provides a speedup of about 250 times over the pairwise transformer baseline ~\cite{WebQA21}  and also is a much lighter model requiring 2 GB gpu memory. Thus our model is more suited for large scale retrieval systems.

\begin{table}[b]
    \centering
    \vline
    \begin{tabular}{c|c|c|c}
        \hline
        \textbf{Question Category} & Train & Val & F1 Score \\
        \hline
        YesNo & 6492 & 828  & 0.698 \\
        \hline
        Number & 1859 & 259 & 0.803 \\ 
        \hline
        Color & 1651 & 179 & 0.735\\
        \hline
        Choose & 3718 & 502 & 0.717\\
        \hline 
        Others & 4743 & 669 & 0.737\\
        \hline
        Shape & 491 & 74 & 0.742\\
    \hline
    \end{tabular}
    \caption{F1 accuracy and Statistics by the Question Category for the visual modality and text modality.}
    \label{tab: question_categories}
\end{table}

\subsection{Qualitative Results}
We analysed a few examples where our model made wrong predictions and other challenging examples where our model was able to predict correctly.

As shown in Figure ~\ref{fig:star} we can see that for the question in row 1, the positive image-based sources
which contain the pillars are captured by the model. Even though pillar is not an explicit class in the image-based model, we see that the model
can identify the correct image sources. Our model was also able to retrieve the other image source though it's caption does not have any lexical overlap with the query. Hence we believe that our model was able to propagate  question-related information across sources. 

\section{Key Insights}
\begin{itemize}
\item Pruning out irrelevant source connections leads to better graph learning as seen in HGNN+SRL model. Full connections to high number negative sources leads to the problem of over-squashing ( phenomenon in which information from the exponentially-growing receptive field is compressed into fixed-length node vectors) and irrelevant information flow.

\item Token level cross attention is more powerful to understand text modality sources as seen in the baseline VLP model which uses token-level features combined with attention. We believe that to be the main reason why we cannot beat the baseline Text F1 as we compress all the text modality information into a single fixed-size vector.

\item GNN training with pre-trained features yields better performance than jointly fine-tuning the pre-trained model as discussed in \ref{Contexrat_Encoding_Module}. 

\item The graph-based architecture helps substantially to improve on Image queries over the baseline VLP model by +4.6 points in F1. However, the performance on text queries degrades. We hypothesize this happens because we compress all text information in one single vector while the baseline uses token level features. However, the HGNN method comes close to the baseline Text F1 despite sub-optimal. We believe combining that graph structure with features used in our best model will beat the baseline on text queries also. We leave that for future investigation.

\end{itemize}
\section{Future Directions}
We further plan to leverae  Graph information flow paths for answer generation might be helpful. 
Answers can be selected either from entities in the constructed entity graph or   from spans in documents by fusing entity representations back into token-level document representation.
\label{conclusion}

\bibliography{main}
\bibliographystyle{icml2021}




\end{document}